%% file: main.tex
\documentclass[10pt,twocolumn,letterpaper]{article}

\usepackage[T1]{fontenc}
\usepackage[margin=0.75in,columnsep=0.25in]{geometry}
\usepackage{times}
\usepackage{graphicx}
\usepackage{amsmath,amssymb}
\usepackage{array,booktabs,multirow}
\usepackage[numbers,sort&compress]{natbib}
\usepackage{xcolor}
\usepackage{microtype}
\usepackage{balance}
\usepackage[font=small,labelfont=bf]{caption}
\usepackage[hyphens]{url}
\usepackage[unicode,breaklinks,hidelinks]{hyperref}

\hypersetup{
  pdftitle={Seeing What Still Holds: How Waiting Control Changes Cooperative Recovery},
  pdfauthor={Kangjie Huang, Yucheng Xin, Xinchun Wang, Dong Liu, Yan Fan, Shuming Liu}
}

\graphicspath{{figures/}}
\newcommand{\framework}{AuditEmbodied}
\newcommand{\safe}{\mathrm{prot}}
\newcommand{\ind}{\mathbb{I}}

\title{\textbf{Seeing What Still Holds: How Waiting Control\\
Changes Cooperative Recovery}}

\author{
  Kangjie Huang\thanks{First and corresponding author.\\
  Email: \href{mailto:202411000534@stu.sdnu.edu.cn}{\texttt{202411000534@stu.sdnu.edu.cn}}.}
  \quad Yucheng Xin\thanks{Yucheng Xin and Xinchun Wang contributed equally as co-second authors.}
  \quad Xinchun Wang\footnotemark[2]
  \\[0.4em]
  Dong Liu\thanks{Dong Liu and Yan Fan contributed equally as co-third authors.}
  \quad Yan Fan\footnotemark[3]
  \quad Shuming Liu
  \\[0.8em]
  {\normalsize School of Computer Science and Artificial Intelligence,}\\
  {\normalsize Shandong Normal University}
}
\date{}

\begin{document}
\raggedbottom
\maketitle

\input{sections/abstract}
\input{sections/introduction}
\input{sections/related}
\input{sections/method}
\input{sections/experiments}

\input{sections/discussion}

\balance
{\footnotesize\raggedright
\bibliographystyle{compactnat}
\bibliography{references}
}

\end{document}

%% file: sections/abstract.tex
\begin{abstract}
Changing the holding reference during a fixed-duration wait can reverse the ranking of visual recovery methods. We introduce \framework, a paired execution and trajectory-evaluation framework that independently varies holding references and waiting duration while keeping recovery executors frozen. A 960-execution factorial on 20 initialization seeds compares CAD-feedback and landmark-based recovery from identical post-wait physical states within each holding and waiting condition. Protected completion requires successful recovery under the declared execution constraints. At a 1\,s wait with registered wrist references, their protected-completion counts change from 6/40 versus 19/40 under registered holding to 34/40 versus 22/40 under position-only holding. Trajectory analysis localizes the main separation to grasp formation: small observation-time target residuals precede large tracking errors. Joint diagnostics and controlled interventions characterize persistent tracking constraints. On fresh seeds, gradual position reanchoring improves CAD recovery's long-wait completion by 42.5 percentage points and reduces mean complete recovery time by 6.06\,s relative to registered holding. Additional evaluations with pretrained policies, handover, and a mass-matched geometry characterize holding-reference sensitivity across executors and physical regimes. These findings establish holding reference as an explicit evaluation condition for comparing visual recovery performance.
\end{abstract}

%% file: sections/introduction.tex
\section{Introduction}

Visual recovery begins in a physical world that continues to evolve during perception and decision making. After unilateral grasp loss in cooperative lifting, the remaining arm may support a tilted object while awaiting a recovery decision. Its holding reference shapes the configuration from which the robot transfers support to a table and re-establishes a dual grasp. Specifying the waiting duration alone therefore does not fully define the physical conditions under which recovery methods are compared. We ask: \emph{Can changing only the reference tracked during a fixed-duration wait reverse the ranking of visual recovery methods?}

Visual monitoring and corrective policies enable recovery from manipulation errors \cite{zhou2025codeasmonitor,xia2025phoenix,zhao2026flare}. Real-time action execution addresses inference latency \cite{black2025rtc}, while intervention-based evaluation reveals policy-dependent responses to pausing \cite{zhou2026liberopro2}. We study the Cartesian reference followed during that interval, independently varying holding and waiting duration while keeping each recovery executor frozen.

We introduce \framework, a framework for paired physical execution and independent trajectory evaluation. \emph{Registered holding} tracks the preceding commanded pose; \emph{position-only holding} reanchors the position target to the measured waiting-entry position while preserving the wrist target. Both use the same low-level feedback and gripper commands. Comparisons across holding conditions share an identical fault prefix. Within each holding and duration condition, the frozen executors also start from identical post-wait physical states.

\begin{figure*}[!t]
\centering
\includegraphics[width=\textwidth]{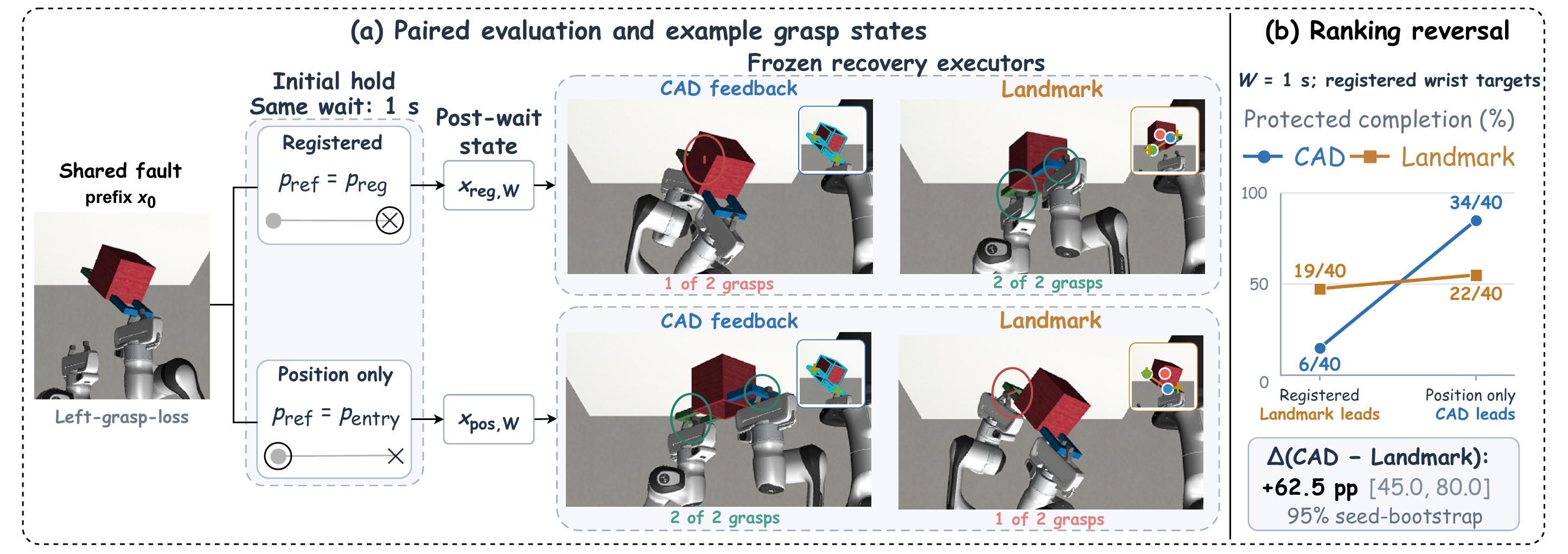}
\caption{
\textbf{Waiting targets reverse the recovery-method ranking.}
(a) Registered and position-only holding start from a shared fault prefix and track different position references. Within each holding condition, frozen CAD-feedback and landmark executors share the same post-wait physical state.
(b) At a fixed 1\,s wait with unchanged wrist references, their protected-completion ranking reverses. Each condition contains 40 fault executions from 20 paired seeds.
}
\label{fig:waiting-execution}
\end{figure*}

A 960-execution factorial crosses two visual executors, CAD feedback using RGB silhouette fitting and landmark recovery using color-centroid triangulation, with four position/wrist holding conditions and two waiting durations on 20 seeds. We evaluate protected completion, defined as successful recovery that also satisfies the declared execution constraints. At a 1\,s wait with registered wrist references, CAD feedback achieves protected completion in 6/40 fault executions, compared with 19/40 for landmark recovery. Reanchoring only the holding position changes these counts to 34/40 and 22/40, respectively, reversing their ranking (Fig.~\ref{fig:waiting-execution}).

Action-checked target reconstruction localizes the main separation to grasp formation: small target residuals at the last visual update precede sharply different tracking errors. Controlled execution studies further characterize persistent tracking constraints and show that placement scheduling modulates holding sensitivity. These findings motivate gradual position reanchoring. On 20 fresh seeds, this improves CAD recovery's long-wait completion rate by 42.5 percentage points and reduces mean complete recovery time by 6.06\,s relative to registered holding.

Our contributions are threefold: (i) we demonstrate through paired execution that changing only the holding reference at a fixed waiting duration can reverse the ranking of frozen visual recovery methods; (ii) we localize the main performance separation to grasp formation, characterize persistent tracking constraints, and validate gradual position reanchoring that improves long-wait completion and reduces complete recovery time on fresh seeds; and (iii) we introduce a reproducible evaluation protocol that combines shared fault prefixes, matched post-wait states within each holding condition, and independent trajectory auditing to jointly assess completion, path protection, and complete costs including failed executions and retries.

%% file: sections/related.tex
\section{Related Work}

\paragraph{Visual failure monitoring.}
REFLECT and AHA diagnose execution failures \cite{liu2023reflect,duan2025aha}. Code-as-Monitor expresses task constraints as visual monitors \cite{zhou2025codeasmonitor}, while RC-NF provides anomaly signals for rollback or replanning \cite{zhou2026rcnf}. We study the physical evolution during the wait between failure detection and recovery.

\paragraph{Recovery policies and skill composition.}
Phoenix translates semantic reflection into motion-conditioned correction \cite{xia2025phoenix}. FLARE combines retry-capable policies with monitor-triggered reset skills \cite{zhao2026flare}, while REACH uses detected failures to guide diffusion policies through negative action prompts \cite{ke2026reach}. RecoveryChaining learns local policies that restore nominal execution \cite{vats2024recoverychaining}, and AgentChord compiles anticipatory recovery branches into task graphs \cite{xu2026agentchord}. We keep recovery executors fixed to isolate the influence of preceding holding control.

\paragraph{Latency and visual control.}
Real-Time Chunking overlaps inference and execution while preserving committed actions \cite{black2025rtc}. Controlled comparisons of asynchronous VLA inference show delay- and benchmark-dependent performance \cite{agouzoul2026asynchronous}. Visual servoing uses image features or reconstructed geometry for feedback control \cite{chaumette2006visual}. We examine how the reference tracked during a physical wait affects subsequent recovery.

\paragraph{Recovery evaluation.}
RoboRecover compares policies from intermediate deviation states reconstructed by action-prefix replay \cite{li2026roborecover}. LIBERO-Pro2.0 compares Hold, Retry, and Reset against normal continuation, revealing policy-dependent responses to pausing \cite{zhou2026liberopro2}. We vary the Cartesian holding reference at fixed waiting durations. Comparisons share fault prefixes across holding conditions and match post-wait physical states across frozen executors within each condition, revealing reference-dependent method rankings.

\paragraph{Trajectory-level evaluation.}
SafeVLA-Bench separates task success from task-aware safety \cite{fan2026safevlabench}, while SafeManip tracks temporal violations and recovery events \cite{huang2026safemanip}. Our independent auditing covers waiting and recovery, checking skill entry, support obligations, and task completion while accounting for failed executions and retries in complete physical costs.

%% file: sections/method.tex
\section{Paired Recovery Evaluation}
\label{sec:framework}

\framework\ evaluates how initial holding references and waiting
duration affect the performance and relative ranking of frozen visual
recovery executors through paired execution and independent evaluation.

\subsection{Problem and Paired Execution}
\label{sec:problem}

Let $x_0$ contain the physical state and permitted history at post-fault
waiting entry. A frozen executor $\pi$, initial holding mode $b$, and
wait $W$ generate the continuous physical trajectory
$\tau(\pi,b,W;x_0)$. The executor uses history $h_t$ of available RGB
images, robot measurements, registered references, and executed commands.
CAD models, calibration, task colors, and setup images are declared
priors where used; evaluator-only object states and contacts never
enter control.

For each initialization and fault or healthy-control scenario,
all $(b,W)$ conditions share $x_0$. Within each core lifting condition,
executors additionally share the post-wait physical state, while
retaining their own perception, observation schedules, and control.
Snapshots and action--observation histories verify both pairing levels;
states across different holding interventions need not match.
Geometry initialization ends before initial grasp closing; later faults
use opening, withdrawal, differential commands, or transient wrenches.
After initial closing, simulation proceeds through faults, waiting,
observation services, recovery, retries, and final verification without
resets, pose edits, or contact anchors.

\subsection{Holding Interventions}

The 20\,Hz Cartesian pose controller executes
$N_W=\lceil W/\delta\rceil$ initial holding steps, with
$\delta=0.05$\,s and core waits $W\in\{0.05,1.00\}$\,s,
excluding subsequent observation-service waits.
For each arm, $\mathbf p\in\mathbb R^3$ and $R\in\mathrm{SO}(3)$
denote end-effector position and orientation in the controller frame.
The registered pose $(\mathbf p_{\mathrm{reg}},R_{\mathrm{reg}})$ is
the last commanded target before waiting; the entry pose
$(\mathbf p_{\mathrm{ent}},R_{\mathrm{ent}})$ is measured once at
waiting entry and held fixed. Suppressing arm indices, the four modes are:

\begin{center}
\small
\setlength{\tabcolsep}{4pt}
\begin{tabular}{@{}lll@{}}
\toprule
Holding mode $b$ & Position target & Orientation target \\
\midrule
Registered
& $\mathbf p_{\mathrm{reg}}$
& $R_{\mathrm{reg}}$ \\
Position only
& $\mathbf p_{\mathrm{ent}}$
& $R_{\mathrm{reg}}$ \\
Rotation only
& $\mathbf p_{\mathrm{reg}}$
& $R_{\mathrm{ent}}$ \\
Position + rotation
& $\mathbf p_{\mathrm{ent}}$
& $R_{\mathrm{ent}}$ \\
\bottomrule
\end{tabular}
\end{center}

Both arms use their own references with unchanged feedback gains and
preceding gripper commands. The intervention affects only the initial
hold; subsequent recovery follows the frozen executor.

\paragraph{Gradual reanchoring.}
A separate smooth-position intervention uses a sampled linear ramp
with $t_{\mathrm{ramp}}=0.25$\,s. At the start of interval
$k=0,\ldots,N_W-1$,
\begin{equation}
\label{eq:smooth-holding}
\begin{aligned}
\mathbf p_k^{\mathrm{ref}}
&=(1-\alpha_k)\mathbf p_{\mathrm{reg}}
  +\alpha_k\mathbf p_{\mathrm{ent}},\\
\alpha_k
&=\min\!\left(1,\frac{k\delta}{t_{\mathrm{ramp}}}\right).
\end{aligned}
\end{equation}
The 50\,ms short hold uses only $k=0$ and therefore retains the
registered position throughout. When $W>0.25$\,s, the reference equals
$\mathbf p_{\mathrm{ent}}$ from $k=5$ onward; this specifies reference
timing, not physical arrival. Registered wrist targets, gripper commands,
gains, and subsequent CAD recovery remain unchanged.

\subsection{Frozen Executors and Timing}

Both core executors share four calibrated RGB views, robot-measurement
permissions, gains, and gripper-force limits.
\emph{CAD feedback} fits colored silhouettes to a registered CAD model,
visually updates two lowering movements, retargets pickup positions and
orientations after table settling, refreshes before descent, recenters
on the table before lifting, and levels after lifting.
\emph{Landmark recovery} triangulates handle-color centroids and
transforms reference grasps by minimum-rotation alignment of the
reference/current handle lines and midpoint translation.
It uses fixed touchdown, release, and clearance phases, refreshes
during segmented approach, and levels on the table before lifting.
Both use pose feedback within phases and allow at most one
encoder-triggered pickup retry. Perception, observation schedules,
and later control are frozen before testing: this compares complete
executors, not perception alone or equal observation budgets.
The auxiliary \emph{Target} uses CAD-based pickup-position correction
with fixed wrist references and phase lengths.

\paragraph{Physical service timing.}
Core reference acquisition and renderer warmup precede $x_0$.
For either core executor, online acquisition-and-estimation service $j$
incurs $d_j=\delta\lceil\ell_j/\delta\rceil$ of physical holding for
measured wall-clock latency $\ell_j$.
Service holds use the executor's original rule, not initial mode $b$.
Actual services and control overruns remain in the physical trajectory;
frozen code does not imply identical latency.
Target retains its two-second minimum inspection slot with charged
padding. Handover uses minimum initial decision windows and
measured-entry wrist references. Released policies retain native
synchronous queries that advance no simulator physics; inserted waits
execute physically, with control duration and query latency reported
separately from lifting costs.

\subsection{Trajectory Evaluation and Cost}

The independent evaluator reconstructs realized skill admission
$V^{\mathrm{eval}}$, raw terminal completion $C$, and path protection
$P$ from physical records and skill boundaries. Protected completion is
\begin{equation}
\label{eq:protected-completion}
Y_{\safe}(\pi,b,W;x_0)
=\ind[V^{\mathrm{eval}}=1]\ind[C=1]\ind[P=1].
\end{equation}
Admission requires all applicable predicates to hold at actual skill
entry after waiting, including continuations and nested calls.
Protection covers shared guards and applicable support-role and
interface obligations.

\paragraph{Trajectory obligations.}
Shared guards reject inter-arm penetration above 10\,mm or intervals
longer than 100\,ms with neither a registered two-finger grasp nor
table contact. Core lifting uses one global call, with internal pickup
retries and no keeper prerequisite. Global restart permits
table-supported release but requires a verified table regrasp cycle.
Local recovery preserves its designated keeper grasp; handover
maintains the active carrier and establishes receiver support before
giver release.
Evaluation spans initial waiting through final verification.
Obligations apply over their declared intervals; violations accumulate
across waits, retries, and calls, and are not erased by later success.
Primary $Y_{500}$ instantiates Eq.~\ref{eq:protected-completion} with
shared guards sampled at 500\,Hz; admission, role, interface, and
terminal tests retain 20\,Hz definitions. Historical cohorts retain
their declared evaluation conventions.

\paragraph{Complete physical cost.}
For simulator times $t_0$ at waiting entry and $t_{\mathrm{end}}$ after
final verification,
\begin{equation}
\label{eq:complete-cost}
\begin{aligned}
T&\equiv t_{\mathrm{end}}-t_0\\
 &=T_{\mathrm{blocking\ wait}}+\sum_iT_{\mathrm{skill},i}\\
 &\quad+T_{\mathrm{release}}+T_{\mathrm{final\ hold}}.
\end{aligned}
\end{equation}
The terms partition elapsed physical time without overlap. Blocking wait
includes initial holding, services, padding, and overruns; $i$ indexes
skill-execution segments, including retries. The remaining terms cover
a separate task-release phase, when present, and terminal verification.
Release actions internal to a skill remain in that skill's cost.
Acquisition, transfer, queueing, and inference count when blocking
continuation; genuinely overlapped computation is not charged twice.
Complete-time means include every path reaching $x_0$, successful or
failed, with all retries. Conditional costs use pairs protected under
both alternatives and are reported separately. Initialization failures
remain in binary denominators, but their post-fault time is undefined,
not zero.

\paragraph{Comparison of interest.}
At fixed $W$, paired comparisons assess reference-dependent executor
rankings. For $Y_{500}$ and $T$, the primary contrast compares the
measured-entry-minus-registered position effect between short and long
waits, averaged over wrist settings, and then between executors
(Eq.~\ref{eq:position-wait}).

%% file: sections/experiments.tex
\section{Experiments}
\label{sec:experiments}

\subsection{Tasks and Evaluation Protocol}

Container recovery and handover use two Panda arms with 20\,Hz
operational-space control (OSC), robosuite~1.5.2, and MuJoCo~3.3.7.
Core lifting uses a 1.818\,kg object and 40\,N gripper limits.
Completion requires both handle grasps, 0.10\,m clearance, tilt within
$15^\circ$, handle-height difference below 0.03\,m, and bounded relative
motion for one second. Handover requires receiver support before giver
release. Each lifting seed supplies left/right faults and a healthy
control; placement varies while other physical settings remain fixed
within each cohort.

Executors are frozen after separate development. Independent cohorts use
fresh seeds; arrival, placement, and geometry extensions reuse matched
seeds. We report $C$, $Y_{500}$, and complete time $T$, retaining failures
in outcome denominators and initialized-path costs. All 95\% intervals
resample whole seeds, preserving fault sides and paired conditions.

\subsection{Holding-Dependent Recovery Rankings}
\label{sec:independent-visual}

\paragraph{Factorial design.}
Twenty seeds cross two executors, four position/wrist references,
two waits ($S=0.05$\,s, $L=1.00$\,s), and left/right faults or healthy
controls: 960 executions. All 60 pre-hold scene groups and 480
within-condition post-wait executor pairs match exactly.
Let $\bar Y^e_{p,r,w}$ denote executor $e$'s fault protection rate,
with position/wrist indicators $p,r\in\{0,1\}$ for registered/measured
references. The prespecified position-by-wait interaction averages
across wrist settings:
\begin{equation}
\label{eq:position-wait}
\begin{aligned}
\Delta_p^e=\frac12\sum_{r=0}^{1}\Big[
&(\bar Y^e_{1,r,L}-\bar Y^e_{0,r,L})\\
-&(\bar Y^e_{1,r,S}-\bar Y^e_{0,r,S})\Big].
\end{aligned}
\end{equation}
We compare $\Delta_p^{\mathrm{CAD}}-\Delta_p^{\mathrm{landmark}}$
and apply the same contrast to all-attempt complete time.

\begin{table*}[!t]
\caption{\textbf{Shared-prefix recovery comparison.}
$S/L$: 0.05/1.00\,s initial waiting; 40 faults per cell from 20 paired seeds.
Healthy pools 20 controls per wait. Complete time includes all fault paths,
failures, visual services, and retries.}
\label{tab:paired-holding-main}
\centering
\small
\setlength{\tabcolsep}{5pt}
\begin{tabular*}{\textwidth}{@{\extracolsep{\fill}}llccccc@{}}
\toprule
\multirow{2}{*}{Executor} &
\multirow{2}{*}{Hold target} &
\multicolumn{2}{c}{Protected completion $Y_{500}$} &
\multirow{2}{*}{Healthy} &
\multicolumn{2}{c}{Complete time (s)} \\
\cmidrule(lr){3-4}\cmidrule(lr){6-7}
& & $S$ & $L$ & & $S$ & $L$ \\
\midrule
CAD feedback & Registered
& 36/40 & 6/40 & 40/40 & 27.01 & 35.58 \\
CAD feedback & Position only
& 34/40 & 34/40 & 40/40 & 26.58 & 27.97 \\
CAD feedback & Rotation only
& 35/40 & 9/40 & 40/40 & 27.03 & 34.41 \\
CAD feedback & Position + rotation
& 36/40 & 36/40 & 40/40 & 27.00 & 28.21 \\
\midrule
Landmark & Registered
& 21/40 & 19/40 & 38/40 & 30.74 & 31.67 \\
Landmark & Position only
& 22/40 & 22/40 & 38/40 & 31.05 & 32.47 \\
Landmark & Rotation only
& 20/40 & 19/40 & 38/40 & 30.66 & 31.87 \\
Landmark & Position + rotation
& 20/40 & 22/40 & 39/40 & 31.04 & 32.27 \\
\bottomrule
\end{tabular*}
\end{table*}

\paragraph{Holding reference reverses the ranking.}
At 1\,s with registered wrists, CAD/landmark protection changes from
6/40 versus 19/40 under registered holding to 34/40 versus 22/40
under position-only holding (Table~\ref{tab:paired-holding-main}).
The paired CAD-minus-landmark gap changes from $-32.5$ [$-52.5,-12.5$]
to $+30.0$ [15.0,45.0] percentage points: a $+62.5$ [45.0,80.0]-point
reversal at fixed delay and implementations. This post-hoc
registered-wrist contrast quantifies the reversal;
Eq.~\ref{eq:position-wait} tests the prespecified wrist-averaged interaction.

\paragraph{Executor-dependent waiting response.}
The position-by-wait interactions are $70.0$ points for CAD and $6.25$
for landmark, giving the primary difference of $63.75$ [47.5,78.75] points.
At 1\,s, position-only holding reduces CAD retries from 37 to 3 and
complete time from 35.58 to 27.97\,s. Rotation-only protection is 9/40,
concentrating the large response in position. In all four CAD
registered/position-only cells, $C=Y_{500}$, linking protection gains
directly to terminal recovery.

\subsection{Execution-Stage Diagnosis and Mitigation}
\label{sec:holding-components}

\paragraph{Grasp formation localizes the separation.}
\label{sec:mechanism-chain}
First-closing targets are independently checked against recorded actions
for all 960 trajectories. Let $\mathbf g$, $\mathbf p$, and $\mathbf h$
denote grasp target, end-effector position, and handle position;
$o$ and $c$ denote the last valid target observation and end of first
closing. The closure-error vector satisfies
\begin{equation}
\label{eq:error-chain}
\mathbf p_c-\mathbf h_c
=(\mathbf g-\mathbf h_o)-(\mathbf h_c-\mathbf h_o)
 +(\mathbf p_c-\mathbf g).
\end{equation}
The terms are observation-time target residual, handle drift, and
tracking error. The target residual includes the registered grasp offset;
vectors are combined before taking norms.

\begin{figure*}[!t]
\centering
\includegraphics[width=\textwidth]{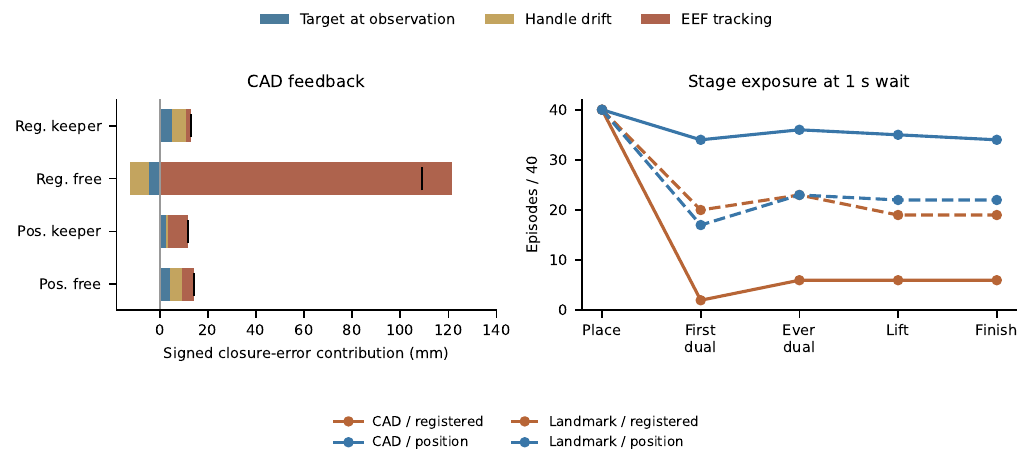}
\caption{\textbf{Error decomposition and recovery stages.}
Each 1\,s, registered-wrist condition retains all 40 faults.
Signed projections sum to closure error; black ticks show its mean norm.
Stage counts include later retries.}
\label{fig:mechanism-chain}
\end{figure*}

At 1\,s with registered wrists, CAD's registered/position-only free-arm
target residuals are 7.1/5.4\,mm, while closing-end tracking errors are
126.8/13.5\,mm. Both conditions place 40/40 objects; first dual grasps
form in 2/40 versus 34/40 (Fig.~\ref{fig:mechanism-chain}).
Tracking errors already differ before closing (128.5/25.9\,mm), locating
the separation before the first closing command. Landmark retains
approximately 64\,mm tracking error and 23/40 eventual dual grasps under
either reference, showing an executor-specific response.

\paragraph{Holding effects persist with arrival adaptation.}
A matched 240-execution CAD variant retains perception, pose feedback,
placement, wrist targets, and evaluation predicates. After each descent,
it tracks for up to 40 extra steps (2\,s), seeking both-arm arrival within
12\,mm and $5^\circ$. At arrival or the cap, it follows the original
best-effort closing and retry rules. Under a shared 45\,s complete-time
budget, eight-query ceiling, and two-attempt limit, long-wait protection is 8/40 registered
versus 37/40 position-only. Its registered-wrist interaction remains
$67.5$ [52.5,82.5] points, versus $75.0$ [60.0,90.0] for original CAD.

\paragraph{Reanchoring improves pre-grasp tracking.}
Analysis of the same 240 adaptive trajectories finds long-wait both-arm
arrival in 3/40 versus 36/40 registered/position-only faults, exhaustion
of the two-second allowance in 37/40 versus 4/40, and mean free-arm
pre-close errors of 100.9/12.6\,mm. Position tolerances alone also pass
in 3/40 versus 36/40, locating the separation in positional tracking.
The free arm lies within 0.02\,rad of a joint bound in 15/40 versus 2/40
paths; normalized translation actions clip on 29.4\% versus 3.0\%
of first-approach steps. Fourteen near-bound registered paths fail,
alongside 18 outside that band. These diagnostics localize pre-grasp
tracking constraints, with joint proximity identifying one affected subset.

\paragraph{Placement modulates the effect.}
\label{sec:placement-main}
A predeclared executor-difference criterion triggers a matched
240-execution probe replacing CAD's two visually estimated lowering
movements with 0.09\,m, 30-step strokes. Observations, charged services,
and later recovery remain unchanged. The registered-wrist interaction
falls from $75.0$ to $20.0$ points (paired change $-55.0$
[$-72.5,-37.5$]), while short-wait registered protection falls from
36/40 to 21/40. This intervention identifies placement scheduling as a
modulator of both holding sensitivity and absolute recovery capability.

\paragraph{Gradual reanchoring retains the benefit.}
Twenty fresh seeds yield 360 executions under registered,
immediate-position, and 0.25\,s smooth-position holding. Long-wait
protection is 8/40, 24/40, and 25/40, respectively, with $C=Y_{500}$.
Smooth holding recovers 17 paired failures while retaining every
registered-hold success: $+42.5$ [27.5,57.5] points. Retries fall from
34 to 7 and complete time from 34.94 to 28.88\,s, a 6.06\,s [4.70,7.40] reduction.
Healthy protection is 119/120. Gradual reanchoring thus improves recovery
and complete time relative to registered holding, with the recovery
executor unchanged.

\subsection{Independent Waiting-Dose Validation}
\label{sec:wait-hold}

\begin{figure*}[!t]
\centering
\includegraphics[width=\textwidth]{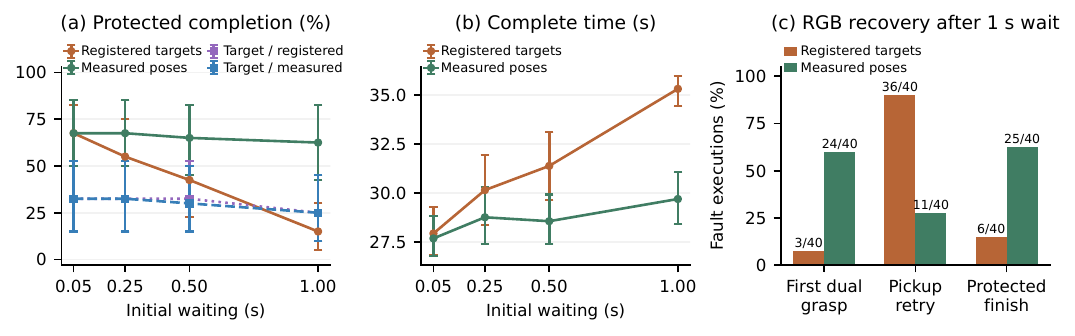}
\caption{\textbf{Independent waiting-dose validation.}
Twenty seeds yield 800 executions; error bars resample whole seeds,
retaining both fault sides. Complete time includes failures, services,
retries, and verification. All endpoint healthy controls protect.}
\label{fig:waiting-overview}
\end{figure*}

Twenty fresh seeds yield 800 executions across four waits
(0.05, 0.25, 0.50, 1.00\,s), CAD/Target executors, and registered/full
measured-pose holding, which reanchors both position and wrist.
CAD protection falls from 27/40 to 6/40 as registered waiting increases,
whereas measured-pose holding maintains 25--27/40
(Fig.~\ref{fig:waiting-overview}); the interaction is
$47.5$ [27.5,67.5] points. At 1\,s, complete time falls from 35.33 to
29.69\,s and CAD's protection gap over Target changes from $-10.0$ to
$+37.5$ points. Fresh seeds and intermediate waits corroborate the
reference-dependent delay response, with measured-pose holding improving
both protection and complete recovery time.

\subsection{Scope across Policies and Physical Regimes}
\label{sec:public-policy}

These cohorts characterize sensitivity under their native tasks and
outcome definitions (Table~\ref{tab:holding-scope}).

\paragraph{Released pretrained visual policies.}
Frozen Diffusion Policy CNN and LSTM--GMM checkpoints
\cite{chi2023diffusion} retain native Transport observations,
700-action horizons, and completion predicate $C$.
Twenty fresh seeds yield 360 executions, with physical state, RGB
history, and RNG/recurrent state matched within each policy before
\emph{extra} $S/L$ waits at action boundaries 32/272.
Native synchronous queries pause simulator physics; control time and
batch-query latency are reported separately from lifting's
physical-service costs. Unpaused completion is 19/20 and 10/20,
respectively. Observed position-by-wait interactions range from $-10.0$
to $+5.0$ points, with all four 95\% intervals including zero
(Table~\ref{tab:public-policy-main}).

\begin{table*}[!t]
\caption{\textbf{Holding interventions on released policies.}
Native Transport completion $C$ uses 20 seeds per cell.
$S/L$: 0.05/1.00\,s extra physical waiting. Interactions are within-policy,
within-boundary position-by-wait contrasts in percentage points;
95\% intervals resample whole seeds.}
\label{tab:public-policy-main}
\centering
\small
\setlength{\tabcolsep}{5pt}
\begin{tabular*}{\textwidth}{@{\extracolsep{\fill}}lcccccc@{}}
\toprule
\multirow{2}{*}{Policy} &
\multirow{2}{*}{Boundary} &
\multicolumn{2}{c}{Registered holding} &
\multicolumn{2}{c}{Position-only holding} &
\multirow{2}{*}{Interaction [95\% CI]} \\
\cmidrule(lr){3-4}\cmidrule(lr){5-6}
& & $S$ & $L$ & $S$ & $L$ & \\
\midrule
Diffusion Policy & 32
& 18/20 & 19/20 & 17/20 & 19/20
& $+5.0$ [0.0,15.0] \\
Diffusion Policy & 272
& 19/20 & 19/20 & 19/20 & 19/20
& $0.0$ [0.0,0.0] \\
\addlinespace[2pt]
LSTM--GMM & 32
& 9/20 & 11/20 & 13/20 & 13/20
& $-10.0$ [-40.0,25.0] \\
LSTM--GMM & 272
& 8/20 & 8/20 & 7/20 & 8/20
& $+5.0$ [-15.0,25.0] \\
\bottomrule
\end{tabular*}
\end{table*}

\paragraph{Explicit grasp-loss recovery.}
A separate 20-seed, 160-execution study injects receiver grasp loss
after dual support. The unchanged Diffusion Policy CNN discards stale
actions, refreshes RGB/robot observations, and resumes.
Native recovery $R$ requires completion after a valid fault prefix.
Both references recover 18/20 assigned faults at $S$;
registered/position-only holding recovers 17/20 versus 18/20 at $L$,
giving a $+5.0$ [0.0,15.0]-point interaction. Two seeds without the
required support event remain in each denominator. This cohort combines
demonstrated grasp-loss recovery with a modest observed holding response.

\paragraph{Cross-task handover.}
\label{sec:handover-holding}
Local receiver regrasp with global retry and direct global restart cross
three position references and $S/L$ minimum waits: 720 attempts on
20 fresh seeds. Scene-wise outcomes match across references:
$C=37/40$ at both waits; $Y_{500}=37/40$ short and 36/40 long.
One initialization-failing seed remains in binary denominators;
costs use 19 initialized seeds. Immediate reanchoring reduces local
long-wait tracking error from 1.00 to 0.54\,mm, with identical observed
completion outcomes.

\paragraph{Mass-matched geometry.}
\label{sec:geometry-transfer}
A shallow container changes dimensions from $14\times14\times14$ to
$18\times11\times8$\,cm and handle spacing from 32 to 26\,cm, retaining
1.818\,kg mass and updating only geometry-dependent CAD/plane constants.
The 240 new position-only executions reuse 240 registered-hold paths on
the same 20 seeds. Target protects 40/40 faults throughout; CAD protects
37/40 registered versus 36/40 position-only at both waits, giving
$0.0$ [$-17.5,17.5$] interaction points. All 240 CAD paths form dual
grasps; incomplete executions are localized to final balance.
Healthy CAD protection is 17/20 and 16/20 registered, versus 17/20 at
both position-held waits.

\begin{table*}[!t]
\caption{\textbf{Scope of holding-reference sensitivity.}
Interactions are within-cohort position-by-wait contrasts in percentage
points. Released CNN uses native recovery $R$; other rows use $Y_{500}$.
Core/adaptive CAD report the registered-wrist slice;
Eq.~\ref{eq:position-wait} averages across wrist settings.}
\label{tab:holding-scope}
\centering
\small
\setlength{\tabcolsep}{4pt}
\begin{tabular*}{\textwidth}{@{\extracolsep{\fill}}lccc>{\raggedright\arraybackslash}p{.28\textwidth}@{}}
\toprule
\multirow{2}{*}{Operating regime} &
\multicolumn{2}{c}{Long-wait outcome} &
\multirow{2}{*}{Interaction (pp)} &
\multirow{2}{*}{Interpretation} \\
\cmidrule(lr){2-3}
& Registered & Position only & & \\
\midrule
Core CAD feedback & 6/40 & 34/40 & $+75.0$ &
Large reference-dependent response \\
Arrival-adaptive CAD & 8/40 & 37/40 & $+67.5$ &
Persists with bounded tracking \\
Released CNN grasp recovery & 17/20 & 18/20 & $+5.0$ &
Capable recovery; small response \\
Mass-matched geometry, CAD & 37/40 & 36/40 & $0.0$ &
Dual grasps formed; balance-limited completion \\
Handover, local recovery & 36/40 & 36/40 & $0.0$ &
Better tracking; unchanged completion \\
\bottomrule
\end{tabular*}
\end{table*}

These within-cohort comparisons connect holding sensitivity to task-stage
outcomes: grasp formation separates the core CAD conditions, final balance
limits shallow-container completion, and handover combines improved
tracking with identical paired outcomes. This motivates evaluating
holding references jointly with the executor and the task stages
governing completion.

\subsection{What Trajectory Auditing Adds}

In an earlier 480-execution holding factorial, five completed paths
from three seeds contain 144--474\,ms registered-support gaps before
successful pickup. A separate recovery-graph cohort completes 60/60
faults and protects 58/60, with shared guards identifying both excluded
paths. Matched supervision experiments identify two additional role-based
exclusions. Together, these checks distinguish terminal completion,
shared support constraints, and skill-specific obligations while
preserving the full execution history.

In the four core CAD registered/position-only cells, $C=Y_{500}$ links
protection gains directly to terminal recovery. Auxiliary cohorts show
how auditing additionally localizes earlier support violations within
successful executions. Together, completion, path compliance, and complete
time characterize recovery capability and execution cost, retaining
waiting, services, failed attempts, and retries; event timing identifies
the affected waiting, transfer, or recovery stages.

%% file: sections/discussion.tex
\section{Discussion}
\label{sec:discussion}

\paragraph{Holding shapes subsequent execution.}
At a fixed delay, the ranking reversal shows that holding control is
part of the recovery system being evaluated. Shared fault prefixes
isolate the initial intervention, and matched post-wait states compare
executors' responses within each condition. The large CAD separation
emerges during grasp formation, where small observation-time target
residuals accompany substantial pre-close tracking errors. Bounded
arrival adaptation preserves a strong reference-dependent response,
with timeouts, joint margins, and action clipping localizing persistent
tracking constraints. The placement probe further shows that earlier
execution stages modulate both sensitivity and completion. Together,
these findings motivate evaluating holding references jointly with
placement and approach timing, following their effects from waiting
through grasp formation.

\paragraph{Practical recovery benefit.}
Gradual position reanchoring raises long-wait completion by
42.5 percentage points and reduces mean complete recovery time by
6.06\,s on 20 fresh seeds, relative to registered holding. Both conditions
use the same 1\,s initial wait, so the time reduction arises in the
subsequent recovery path, alongside a drop in pickup retries from 34
to 7. The intervention uses measured robot positions while preserving
the observer, wrist targets, gains, gripper commands, and later recovery
rules. This provides a concrete way to improve the waiting-to-recovery
transition within the existing controller. Complete-time means retain
services, unsuccessful attempts, and retries across all initialized
fault paths, quantifying elapsed execution cost across the full cohort
alongside the completion gain.

\paragraph{Task-stage dependence.}
Cross-regime experiments connect reference sensitivity to the stages
governing completion. In the mass-matched shallow geometry, all CAD
executions form dual grasps, while final balance determines the remaining
outcomes. Handover combines lower tracking error with unchanged
completion, and released policies show modest observed interactions
under their native tasks. These within-cohort results distinguish
tracking improvement from its effect on the task goal. For recovery
design, they suggest prioritizing the stage that limits completion:
grasp formation in the core lifting regime, and final balance in the
shallow-container regime. Reporting grasp acquisition, support transfer,
and balance alongside aggregate success makes these differences
actionable.

\paragraph{Evaluation practice and scope.}
Recovery evaluations should report position and wrist references,
physical actions during waiting, and whether delay specifies a total
hold or an added pause. Common fault prefixes and matched
within-condition recovery states make reference-dependent rankings
interpretable. Joint reporting of completion, path obligations, stage
outcomes, and complete time captures capability, execution history,
and cost; whole-seed resampling preserves pairing across fault sides
and repeated conditions. Paired contrasts quantify response magnitude;
stage traces localize its emergence. Native synchronous-policy query
latency retains its separate timing convention. The present evidence
covers Panda/OSC simulation, with registered geometry, calibrated
cameras, and task-color priors for the container executors.
Hardware evaluation can apply the same paired protocol to sensing
delay, contact dynamics, and controller design, while additional task
families extend the range of support transitions.

\section{Conclusion}
\label{sec:conclusion}

\framework\ combines paired physical execution and independent
trajectory evaluation to study visual recovery under controlled holding
and waiting. Changing only the initial position reference at a fixed
delay reverses the ranking of two frozen recovery executors. Stage
analysis localizes the large response to grasp formation and tracking
constraints, and gradual reanchoring improves completion and complete
recovery time on fresh seeds. Across operating regimes, the results
make holding reference an explicit condition for recovery evaluation
and a concrete control choice to assess jointly with the executor
and task.